\documentclass[sigconf]{acmart}
\usepackage{multicol}
\usepackage{amsmath}
\usepackage{enumerate}
\usepackage[T1]{fontenc}
\usepackage{mathrsfs}
\usepackage{hyperref}
\usepackage{derivative}
\usepackage{graphicx}
\usepackage{caption}
\usepackage{subcaption}
\usepackage{amsmath}
\usepackage{algorithm}
\usepackage[noend]{algpseudocode}
\usepackage{mathtools}
\graphicspath{ {./images/} }
\usepackage{xcolor}
\PassOptionsToPackage{hyphens}{url}\usepackage{hyperref}
\usepackage{float}

\setcopyright{acmcopyright}
\copyrightyear{2021}
\acmYear{2021}
\acmDOI{}

\acmConference[Responsible AI '21]{Responsible AI 2021: 1st ACM SIGKDD Workshop on Measures and Best Practices for  Responsible AI}{August 2021}{KDD Virtual Conference}
\acmBooktitle{Responsible AI 2021: 1st ACM SIGKDD Workshop on Measures and Best Practices for  Responsible AI, August 2021, KDD Virtual Conference}
\acmPrice{15.00}
\acmISBN{978-1-4503-XXXX-X/18/06}

\newtheorem{innercustomgeneric}{\customgenericname}
\providecommand{\customgenericname}{}
\newcommand{\newcustomtheorem}[2]{%
  \newenvironment{#1}[1]
  {%
   \renewcommand\customgenericname{#2}%
   \renewcommand\theinnercustomgeneric{##1}%
   \innercustomgeneric
  }
  {\endinnercustomgeneric}
}

\newcustomtheorem{customthm}{Theorem}
\newcustomtheorem{customlemma}{Lemma}
\newcustomtheorem{customdef}{Definition}
\newcustomtheorem{customcor}{Corollary}

\makeatletter
\def\BState{\State\hskip-\ALG@thistlm}
\makeatother

\title{PCACE: A Statistical Approach to Ranking Neurons for CNN Interpretability}

\author{S\'{\i}lvia Casacuberta}
\affiliation{%
  \institution{Harvard University}}
\email{scasacubertapuig@college.harvard.edu}

\author{Esra Suel}
\affiliation{%
  \institution{Imperial College London}}
\email{esra.suel@imperial.ac.uk}

\author{Seth Flaxman}
\affiliation{%
  \institution{University of Oxford}}
\email{seth.flaxman@cs.ox.ac.uk}

\begin{document}

\begin{abstract}
    In this paper we introduce a new problem within the growing literature of interpretability for convolution neural networks (CNNs). While previous work has focused on the question of how to visually interpret CNNs, we ask what it is that we care to interpret, that is, which layers and neurons are worth our attention? Due to the vast size of modern deep learning network architectures, automated, quantitative methods are needed to rank the relative importance of neurons so as to provide an answer to this question. We present a new statistical method for ranking the hidden neurons in any convolutional layer of a network. We define importance as the maximal correlation between the activation maps and the class score. We provide different ways in which this method can be used for visualization purposes with MNIST and ImageNet, and show a real-world application of our method to air pollution prediction with street-level images.
\end{abstract}

\maketitle

\section{Introduction}
In recent years, the use of convolutional neural networks (CNNs) has become widespread due to their success at performing tasks such as image classification or speech recognition. CNNs have achieved outstanding results at the ImageNet challenge and popularized the use of such architectures \cite{yosinski}. Remarkably, these same networks that outperform at the ImageNet challenge are also successful when used for other tasks such as object detection on the PASCAL VOC dataset~\cite{zhou}. Large training datasets, powerful GPUs and the implementation of regularization techniques such as dropout or regularization have also helped boost the performance of CNN architectures \cite{zeiler}.

While CNNs continue to excel at countless tasks and competitions, their internal procedures remain a mystery and there is very little insight into how these architectures achieve such outstanding results. We effectively treat these neural networks as black boxes and it is extremely challenging to understand how they operate, due to their complexity and large number of interacting parts. However complex these black boxes might be, it is vital to acquire a deeper understanding of how they work. There are several reasons for encouraging research in this area. From a scientific perspective, understanding how deep neural networks operate is interesting on its own, but mastering their inner mechanisms will also allow us to improve their results and accuracy. Without any insight on how these black boxes work, their development into better models can only be achieved by trial-and-error. From a social perspective, we should not be allowing a system that applies untransparent and unexplainable algorithms to make decisions that govern health care, banking, or politics. Once the decisions taken by deep neural networks shift from innocuously classifying hand-written digits to deciding whether someone has a particular disease or is eligible for a bank loan, it becomes imperative to advocate for a right to explanation \cite{goodman}.

This paper focuses on the following question: how can we \textit{rank} the hidden units of a convolutional layer in order of importance towards the final classification?\footnote{We use the words \textit{unit}, \textit{filter}, \textit{channel}, and \textit{neuron} interchangeably, as there does not appear to be consensus in the literature~\cite{li}.} While research around understanding the inner mechanisms of CNN architecture advances, visualization methods that are targeted directly to each neuron (e.g., feature visualization) are unfeasible to apply and analyze to all of the thousands of neurons in the architecture. Therefore, our goal is to provide a method that identifies the neurons that contribute the most to the final classification. On top of it, any other visualization or explainability method that applies to a particular neuron can be studied only on this small set of ranked neurons or combinations of them, hereby making it practical. We discuss this question thoroughly and propose a novel statistical method called \textit{PCACE} that ranks the hidden units of a convolutional layer according to their relevance towards the final (class) score. The algorithm is based on the Alternating Conditional Expectation algorithm by Breiman and Friedman~\cite{breiman}, which provides the optimal transformations that allow us to maximize the correlation between the activation maps of the neurons with the final (class) score.

We show how to use our statistical method for visualization purposes by using activation maps, CAM, and the activation maximization method. The combination of the statistical algorithm behind PCACE with several visualization methods yields a new procedure to help in the interpretability and explainability of convolutional neural networks. Besides testing our algorithm on the well-known datasets of MNIST and ImageNet, we also provide a real-world use case to air pollution prediction of street-level images. In the case of ImageNet, we replicate our results with both the ResNet-18 and VGG-16 architectures and compare them. We analyze the general features shown by our PCACE algorithm and consider the differences between gradient-based and statistical-based interpretability methods. 

\section{Related Work on Interpretability and Explainability of Neural Networks}\label{sec:relwork}
After the rapid success of deep neural networks, the community has recently started to acknowledge that we have very limited understanding of how these architectures work and how they are able to achieve such remarkable results. Some of the first visualization tools that were proposed were \textit{saliency maps}~\cite{simonyan}, which indicate the areas of the input image that are discriminative and most important with respect to the given class by using the intensity of the pixels. That is, given an input image, the saliency method ranks its pixels based on their influence on the final class score by using derivatives and backpropagation. A similar idea is the \textit{Grad-CAM method}~\cite{selvaraju}, which employs the gradients of any target concept to produce a localization map that highlights the important regions in the image that predict the concept. Grad-CAM can also be combined with existing pixel-space visualizations (Guided Grad-CAM) to achieve visualizations that are both high-resolution and class-discriminative.

%\begin{figure}[ht]
  %\centering
  %{\includegraphics[width=5.5cm]{saliency.png}}
  %{\includegraphics[width=5.5cm]{gradcam.png}}
  %\caption{Example of saliency map (top) and Grad-CAM (bottom) applied to input digit 2 in MNIST.}
%\end{figure}

Another way to study the importance of the pixels in the input image is to \textit{perturb the image} by occluding patches of the image and see how the classification score drops~\cite{zeiler}. Other authors follow a similar idea by directly deleting some parts of the image in order to find the part of the image that makes the final class score drop the most~\cite{fong1}. A bit differently, Koh and Liang use \textit{influence functions} to gain understanding on which training points are more relevant to the final classification without having to retrain the network~\cite{koh}. Their method essentially shows how the model parameters change as we upweight a training point by an infinitesimal amount.

In~\cite{zeiler}, the authors try to understand CNNs backwards by creating a Deconvolutional Network. A \textit{DeConvNet} can be thought as a ConvNet model with the same components but in reverse, so that it maps features to pixels. A similar idea is followed in~\cite{dosovitskiy}, where image representations are studied by inverting them with an up-convolutional neural network. Other work focuses on developing human-friendly concepts to help understand the machine. In~\cite{kim}, \textit{Concept Activation Vectors} are introduced to provide an interpretation of the internal state of a network in human-friendly concepts, and \textit{Deep Dream} and \textit{Lucid} are Google projects that intend to humanize what the hidden layers and neurons see in the input images~\cite{olah}. Their paper develops human-like visualizations for understanding what each neuron is focusing on. In the conclusion, the authors point out that one of the issues that still stands out in network interpretability is finding which units are most meaningful for understanding neural net activations, which is what we study in this paper. Other studies also engage in trying to bridge human concepts and neural networks; for example,~\cite{zhou} investigates how transferable are features in deep neural networks by differentiating between general and specific features learnt by the architecture. It was asked in~\cite{alsallakh} if CNNs learn class hierarchy, and \cite{bau} contains a study of the semantic concepts learnt by the units, such as colors, scenes, and textures.

Other papers bring up criticism to some of the methods we just described. In~\cite{kindermans1}, it is argued that saliency methods lack reliability when the explanation is sensitive to factors that do not contribute to the model prediction, and in \cite{kindermans2} it is shown that DeConvNets and Guided Backpropagation do not produce the theoretically correct explanations for a linear model, and so even less for a multi-layer network with millions of parameters. Finally, in \cite{fong2} and \cite{nguyen}, the authors propose that neurons do not encode single concepts and that they are in fact multifaceted, with some concepts being encoded by a group of neurons rather than by a sole neuron by itself.

As summarized in this section, most papers in the literature focus on qualitatively studying the specific features and concepts that are being learnt in the network, rather than quantifying the importance of each hidden neuron towards the final class score. Our method bridges this gap by focusing on quantifying the relevance of each neuron, beyond providing qualitative information based on visualization methods. An exception in the literature is the recent paper \cite{bau2020}, where the most relevant units for each class are defined by computing which cause the most accuracy loss when removed individually. However, this approach is too computationally costly.

\subsection{Class Activation Mapping}
It was first observed in~\cite{zhou} that CNNs behave as object detectors even though no location information about the central object is provided. In this paper, we provide more evidence supporting this claim. In~\cite{cam} they combine a simple modification of the global average pooling layer with the class activation mapping (CAM) technique to allow for a classification-trained CNN to also be able to localize specific image regions in a single forward-pass. Examples are provided in Sections 5 and 7, where we compare the activation maps of different neurons with the CAM visualization of the same input image.

\subsection{Activation Maximization}
The activation maximization method~\cite{erhan}, instead of highlighting discriminative regions of the input image (as it is the case of saliency maps and CAM), synthesizes an artificial image $x^{*}$ (which we will henceforth call \textit{ideal image}) that maximizes the activation of a target neuron~\cite{zhuwei}:
\begin{equation}
    x^{*}= \textrm{argmax}_{x}\, a_{i, l} (\Theta, x),
\end{equation}
where $\Theta$ denotes the network parameter sets; i.e., the weights and the bias. This is achieved through an iterative process: after initially setting a random image $x_0$, the gradients with respect to $x_0$, that is, %$\pdv{a_{i, l}}{x}$, 
$\partial{a_{i,l}}/\partial x$,
are computed with backpropagation. Each pixel of the initial noisy image is changed iteratively to maximize the activation of the neuron, applying the update
\begin{equation}
    x \longleftarrow x + \gamma \cdot \pdv{a_{i, l}(\Theta, x)}{x},
\end{equation}
where $\gamma$ denotes the gradient ascent step size. The final image $x$ represents the \textit{preferred} input for that neuron.

\begin{figure}[ht]
  \centering
  {\includegraphics[width=0.268\textwidth]{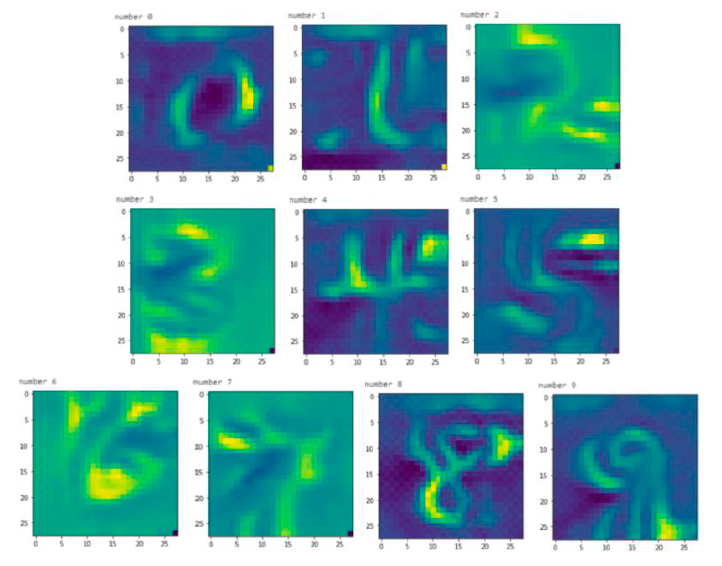}} \hspace{-0.21cm}
  {\includegraphics[width=0.212\textwidth]{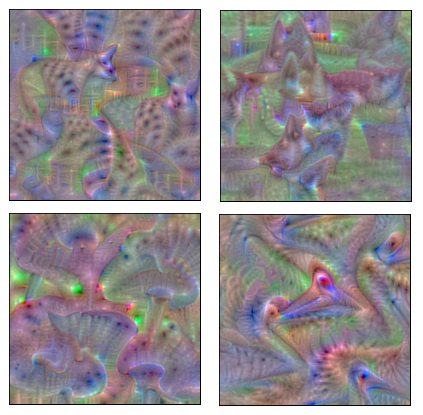}}
  \caption{Ideal images synthesized with the activation maximization method for MNIST (left), and the ideal images for the ImageNet classes \textit{Egyptian cat}, \textit{German shepherd}, \textit{Pelican}, and \textit{Mushroom} in VGG-16 (clockwise at right).}
  \label{fig:ideal}
  \vspace{-0.3cm}
\end{figure}

Note that the activation maximization method allows to both create the ideal image that maximally activates a neuron (Section~6), as well as the ideal image for a particular class (Figure~\ref{fig:ideal}).
Moreover, the activation maximization method uses the unnormalized final class score immediately before the application of the softmax function, to prevent the values from being squeezed between $0$ and~$1$. We will follow the same principle for our PCACE algorithm. In order to make the ideal image more interpretable, the activation maximization method is normally used with regularization methods, such as $\ell_2$ decay or Gaussian blur~\cite{zhuwei}.

\section{Statistical Methods For Interpretability}\label{sec:pcace}
\subsection{Alternating Conditional Expectation}
This section introduces a novel statistical method to rank the hidden units of any neural network in order of importance towards the final class score: PCACE. The name PCACE results from the fusion of PCA (Principal Component Analysis), which we use for dimensionality reduction, and ACE (Alternating Condition Expectation), which we use to compute the maximal correlation coefficient between a dependent variable (the final class score) and multiple independent variables (each entry of the activation matrix of a particular neuron). The units will be ranked on the basis of the strength of their possibly non-linear relationship with the final class score. 
% Therefore, we will rank the units based on the strength of the non-linear relationship to the final class score. 

The Alternating Conditional Expectation (ACE) algorithm~\cite{breiman} is a non-parametric approach for estimating the transformations that lead to the maximal multiple correlation of a response and a set of independent variables in regression and correlation analysis. The iterative algorithm works by estimating optimal transformations in a multiple regression setup,  to find the maximal correlation between multiple independent variables $X_i$ and a dependent variable~$Y$. These transformations minimize the unexplained variance of a linear relationship between the transformed response variable and the sum of the transformed predictor variables~\cite{wang}. Moreover, ACE does not require any assumptions on the response or predictor variables and is entirely automatic, in contrast to more recent methods for nonparametric dependence testing (e.g.,~kernel methods like the Hilbert-Schmidt Independence Criterion \cite{gretton2008kernel} require specifying a kernel and its lengthscale). ACE also contrasts with Generalised Additive Models in that it transforms the response variable. 

As summarized in~\cite{wang}, the general regression model for~$p$ independent variables (predictors) $X_1, X_2, \ldots, X_p$ and a response variable $Y$ is given by
\begin{equation}
    Y = \beta_0 + \sum_{i=1}^p \beta_i X_i + \epsilon,
\end{equation}
where $\beta_0, \ldots, \beta_p$ are regression coefficients that are to be estimated, and $\epsilon$ is the error term. However, this model is not well-suited for CNN architectures, since the activation functions in the network are \textit{not} linear, and so instead we must use a non-linear regression method.
The ACE regression model has the form
\begin{equation}
    \Theta(Y) = \alpha + \sum_{i=1}^p \phi_i(X_i) + \epsilon,
\end{equation}
where $\Theta(Y)$ is a function of the response variable $Y$ with zero mean and unit variance, and $\phi_i(X_i)$ are zero-mean functions of the predictors $X_i$, for $i = 1, \ldots, p$. Then ACE finds the functions that minimize the error variance not explained in the regression, i.e.,
\begin{equation}
    \epsilon^2(\Theta, \phi_1, \ldots, \phi_p) = E\Big\{\Big[\Theta(Y) - \sum_{i=1}^{p} \phi_i(X_i)\Big]^2\Big\},
\end{equation}
with respect to $\Theta$ and $\phi_1, \ldots, \phi_p$. 

In our case, given a fixed neuron in a convolutional layer in the architecture, the independent variables $X_i$ correspond to the entries of the activation map produced from the convolution between the fixed weights of the neuron and the input image. The predictor variable $Y$ corresponds to the final class score of the input image \textit{before} the application of the softmax function, to avoid the compression of the values into the interval $[0, 1]$. Since ACE will assign a number between $-1$ and 1 to each hidden unit, it allows to rank all of the hidden units of the network in a reliable, normalized, and clear way. The higher the ACE value is, the stronger is the relationship between that neuron and the score of a particular class. 

%\begin{figure}[ht]
%\includegraphics[width=8cm]{fotoo.png}
%\caption{Visual explanation of the relationship between the matrix of weights and the activation map of a neuron (cite).}
%\centering
%\end{figure}

\subsection{Applying ACE to CNN Architectures}
We now give a more detailed explanation of how to apply ACE to the CNN setting. Given a trained CNN architecture, we fix one of the convolutional layers, and our goal is to produce a ranking of the $m$ neurons that form that layer.\footnote{As noted in the introduction, PCACE can produce a global ranking of the neurons in the network because the values are normalized. However, in this paper we decide to focus on each layer separately in order to be able to compare the magnitude of the PCACE values across them.} Each of these neurons has a fixed matrices of weights, which has size corresponding to the kernel of the convolutional layer. This matrix of weights produces an activation map whenever an input image is fed into the network and the neuron is then fired by its activation function. We fix a neuron $N_i$ in this convolutional layer, and we fix a set of input images which will be used to compute the maximal correlation between neuron $N_i$ and the final class score of this set of images. 

It can be meaningful to either have this set correspond entirely to training images, entirely to unseen images by the network, or a mix of both. It is also recommended to make this set of pictures be all of the same class, so that we are correlating a neuron with its importance towards the classification of a particular class (for example, in MNIST, we can create a ranking of the neurons for each of the 10 digits separately). However, we note that our algorithm PCACE does not require to be class specific (as we use it in Sections~\ref{sec:mnist} and \ref{sec:imagenet}), and can also be applied in regression tasks (as we do in Section~\ref{sec:airpoll} for spatiotemporal data). Even for classification tasks, it can be interesting to use a set of input images that consists of a weighted mix of all classes, in order to produce a ranking of the overall most correlated neurons.

Each of these neurons produces an activation map when an input image is fed into the network, which is a matrix of size $k_1 \times k_2$. ACE will then be working with $k_1 \cdot k_2$ independent variables $X_i$, where each corresponds to one of the entries in the activation maximization map. We favor this method instead of simpler ones (e.g., those consisting of only taking the mean or the maximum value of the $k_1 \cdot k_2$ activation values in order to have just one independent variable), which neglect the complexity and range of values within the activation map. The response variable $Y$ is given by the final correct class score right before applying the softmax function. For each pass of one of the $n$ input images of the set, we store the $k_1 \cdot k_2$ activation maximization values as a column in a matrix $M$, and the final class score in a vector $v$.

%\begin{figure}[ht]
%\includegraphics[width=8.5cm]{diagram.png}
%\caption{Visual diagram of the procedure behind the PCACE algorithm.}
%\centering
%\end{figure}

At this stage, we cannot input $M$ and $v$ directly into the ACE algorithm. Firstly, the dimensions of both the matrix $M$ and the vector $v$ are normally too large for ACE to be able to compute the maximal correlation coefficient in a reasonable computation time. Secondly, the ACE algorithm halts and outputs \texttt{nan} if values in the matrix $M$ are too small due to division by 0. One way to fix this is to standardize each row of the matrix $M$ (i.e., standardize each of the $k_1 \cdot k_2$ predictor variable vectors): center to the mean and component-wise scale to unit variance. However, the issue might still not be resolved if the standard deviation of any of the $k_1 \cdot k_2$ predictor variables is 0.

Both problems can be solved at once by applying the PCA (Principal Component Analysis) algorithm before inputting $M$ and $v$ into ACE. PCA performs a linear mapping of the data in the original matrix to a lower-dimensional space such that the variance of the data in the low-dimensional representation is maximized by using eigenvalue decomposition~\cite{jolliffe}. After applying PCA, the new matrix $M'$ will have smaller dimensions but will have inherited the maximum possible variance of the original data. Once the dimension of the original matrix $M$ has been reduced, we can now apply ACE in a computationally efficient way. It is important to remark that when using PCA for the purposes of PCACE one needs to reduce the number of rows (the number of predictor variables), and not the number of columns (the number of input images). 

%We will also take the absolute value of the final PCACE algorithm, as we are concerned with the magnitude of the correlation. We remark that very few channels in our experiments presented a negative ACE correlation before taking the absolute value.

We take the absolute value of the final PCACE value, as we are concerned with the magnitude of the correlation. In fact, very few channels in our experiments presented a negative ACE correlation.

\subsection{The PCACE Algorithm}
PCACE is an algorithm that can be effectively used to rank hidden channels for the following reasons:
\begin{enumerate}
    \item It bridges together two effective, reliable, and powerful statistical methods: the Alternating Conditional Expectation (ACE) and the Principal Component Analysis (PCA).
    \item By using PCA on top of ACE, we can significantly reduce the dimensions of the matrix $M$, therefore making the ranking computation time efficient.
    \item Since the number of predictor variables is significantly reduced, we can increment the number of input images and PCACE will still be computationally efficient.
    \item ACE will not encounter \texttt{nan} problems because PCA standardizes the original data.
    \item The output PCACE values are all between $0$ and $1$, %(after we apply the absolute value), 
    which yields a %statistical and 
    standardized method to make comparisons across different neurons and from different layers.
\end{enumerate}

%Figure~5 summarizes the pseudocode of the PCACE algorithm in a generic CNN architecture.

%\begin{figure}[ht]
%\includegraphics[width=8.3cm]{pseudocode.png}
%\caption{Pseudocode of the PCACE algorithm.}
%\centering
%\end{figure}

%\begin{algorithm}
%\end{algorithm}

\section{An Example with MNIST}\label{sec:mnist}
We first provide an example of applying the PCACE algorithm to the MNIST dataset to produce a ranking of the neurons in the first convolutional layer for each of the 10 digits. The size of the activation matrix of the hidden units of the first convolutional layer in our architecture is $26 \times 26$, which means there are 676 predictor variables $X_i$. For each input image, we record the activation value that each of the elements in the activation matrix achieves, store it, and record the final class score. After feeding $m$ input images into the network, we obtain 676 vectors of length $m$ for the predictor variables, and one vector of length $m$ for the response variable. We use 500 input images of each particular digit from the training dataset. After repeating this procedure with all the neurons in the layer for each of the 10 digits, we obtain the final PCACE rankings. 

We observe that many of the channels that PCACE ranks as the most important for each digit are precisely the ones that focus on the surroundings and details of the digit instead of the global shape, which we call \textit{reverse channels}. %Moreover, 
The reverse channels for MNIST are numerically characterized by having a negative weight mean. %The reverse channels 
They appear to activate at the edges, corners, and surroundings of the digit when the input digit is a real image from the training set (or a mean of the images in the training set), but are nearly dead (i.e., activation close to~0) when the input image is the ideal class digit created with the activation maximization method.
%, as shown in Figure~\ref{fig:fail}. 

This indicates a potential shortcoming of the activation maximization method: the artificial image produced by the activation maximization does not activate most of the channels that are ranked by the PCACE algorithm as the most important for the correct classification of the digit. These channels do not focus on the main class object (the digit), and therefore are not fired because the pixels that they identify are not maximized by backpropagation. More broadly, we aim to point out that interpretability methods that are gradient-based (such as CAM or saliency maps) do not necessarily correlate well with those that are statistics-based (such as the correlation method of PCACE). As we further develop in Sections~\ref{sec:imagenet} and \ref{sec:filter}, the channels in CNN architectures encode more information than that explicitly related to the class object, and this complexity is oftentimes lost with gradient-based methods.
    
%\begin{figure}[ht]
%\includegraphics[width=8.5cm]{activationmax.png}
%\caption{Visualization of how the ideal digits created by the activation maximization method fail to fire most of the neurons that are ranked as top 5 in the PCACE algorithm.}
%\centering
%\label{fig:fail}
%\end{figure}

\vspace*{-0.1cm}

\section{ImageNet Results}\label{sec:imagenet}
We try our PCACE algorithm with the more complex ImageNet (1${,}$000) dataset, on both the architectures ResNet-18 and VGG-16. With ResNet-18 we are able to use CAM, and thus we can compare the CAM visualizations with the activation maps of the top PCACE channels. While it is not possible to obtain CAM visualizations with VGG-16~\cite{cam}, the first convolutional layers preserve the size of the input image, which allows for a clearer visualization of the activation maps ($112 \times 112$ for ResNet-18 as opposed to $224 \times 224$ for VGG-16). We apply the PCACE algorithm on all convolutional layers of both architectures for several ImageNet classes, using 300 input images for each class (randomly selected from the ImageNet dataset), and using PCA to reduce by half the size of the matrix that we input to the ACE algorithm. The algorithm yields class-specific rankings of the neurons in each convolutional layer (ranked independently in each layer). 

% [Discuss more about visualizing the intermediate activations]

% Show the locative power of CNNs despite being trained only on labels for classification.

\begin{figure}[ht]
  \centering
  {\includegraphics[width=0.15\textwidth]{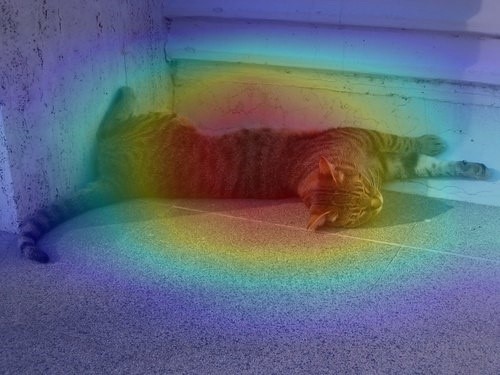}}
  %\hfill 
  \quad
  {\includegraphics[width=0.14\textwidth]{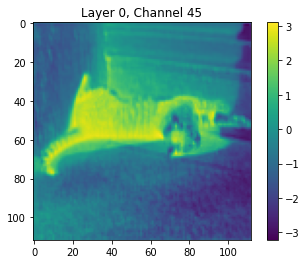}}
  {\includegraphics[width=0.14\textwidth]{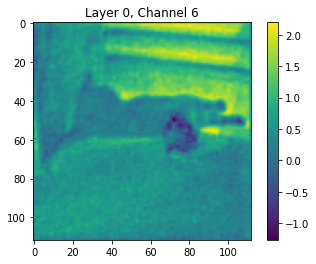}}
  \par\medskip
  {\includegraphics[width=0.15\textwidth]{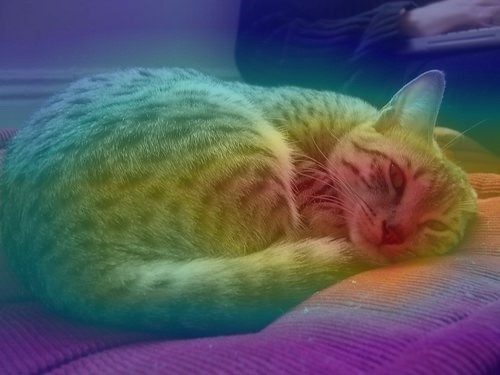}}
  %\hfill 
  \quad
  {\includegraphics[width=0.14\textwidth]{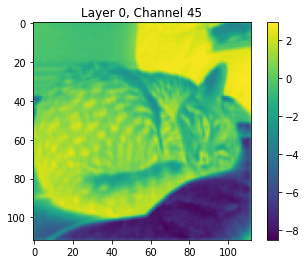}}
  {\includegraphics[width=0.14\textwidth]{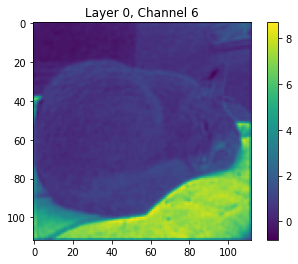}}
  \par\medskip
  {\includegraphics[width=0.15\textwidth]{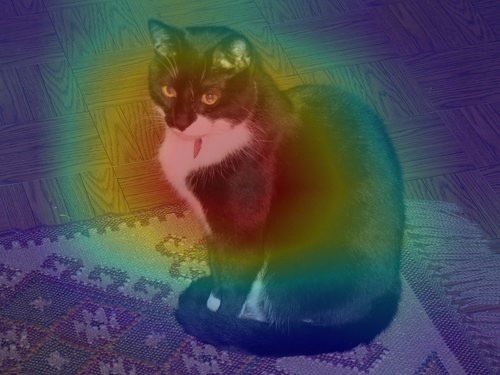}}
  %\hfill 
  \quad
  {\includegraphics[width=0.14\textwidth]{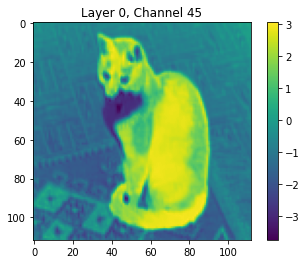}}
  {\includegraphics[width=0.14\textwidth]{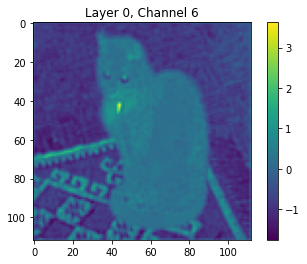}}
  \caption{CAM visualization (left), activation map of the bottom PCACE channel (middle), and activation map of the second highest PCACE channel (rigth) with ResNet-18 for the ImageNet class \textit{Egyptian cat}.}
  \label{fig:CAM}
\end{figure}

\begin{figure}[htbp]
  \centering
  {\hspace{0.1cm}\includegraphics[width=0.14\textwidth]{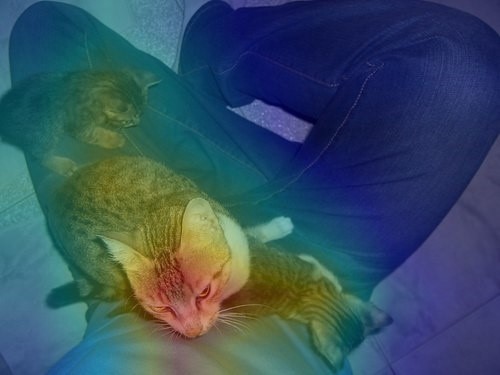}}
  %\hfill
  \quad
  {\includegraphics[width=0.14\textwidth]{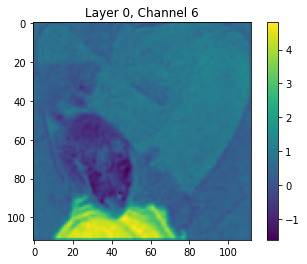}}
  {\includegraphics[width=0.14\textwidth]{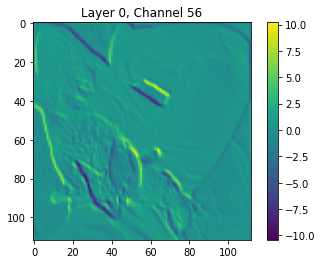}}
  \par\bigskip
  {\hspace{0.4cm}\includegraphics[width=0.10\textwidth]{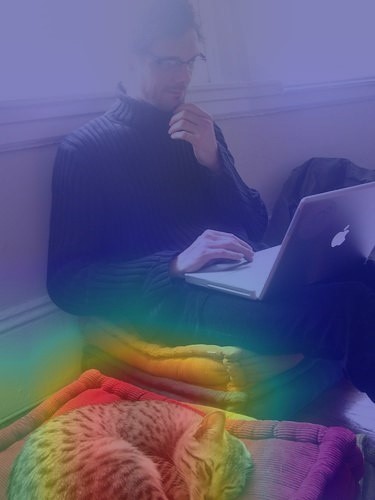}}
  %\hfill
  \quad
  {\includegraphics[width=0.14\textwidth]{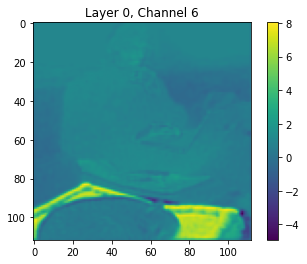}}
  {\includegraphics[width=0.14\textwidth]{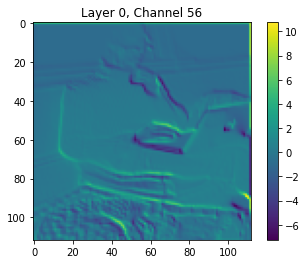}}
  \par\bigskip
  {\hspace{0.1cm}\includegraphics[width=0.14\textwidth]{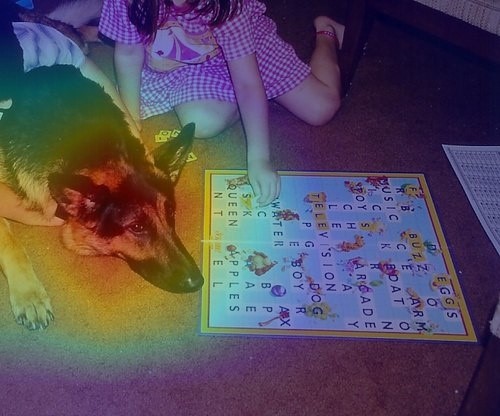}}
  %\hfill
  \quad
  {\includegraphics[width=0.14\textwidth]{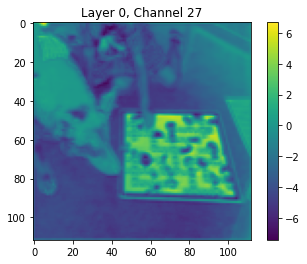}}
  {\includegraphics[width=0.14\textwidth]{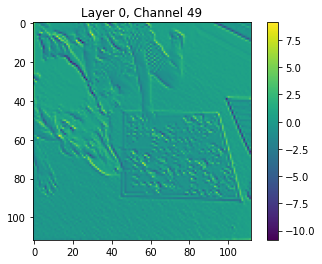}}
  \par\bigskip
  {\includegraphics[width=0.15\textwidth]{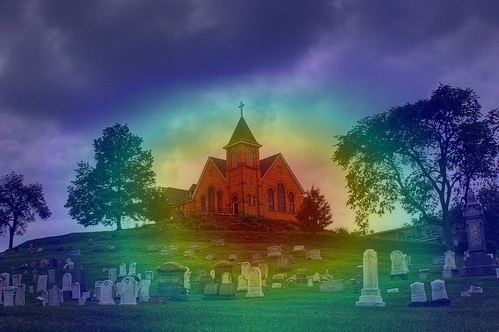}}
  %\hfill
  \quad
  {\includegraphics[width=0.14\textwidth]{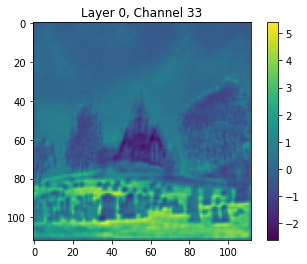}}
  {\includegraphics[width=0.14\textwidth]{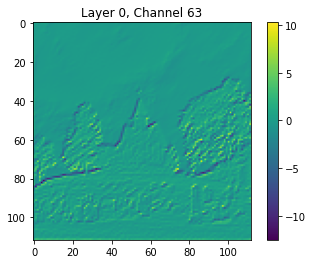}}
  \caption{CAM visualization (left) and the activation maps of two of the top 5 PCACE channels (middle, right) for the ImageNet classes \textit{Egyptian cat}, \textit{German shepherd}, and \textit{Church} in Res\-Net-18.}
  \label{fig:ImageNet}
\end{figure}

We then plot the activation maps of the top PCACE channels (i.e., those that show the highest correlations) and compare them to the CAM visualization of~\cite{cam}. Examples are shown in Figure~\ref{fig:CAM}. %Firstly, 
We support the claim in~\cite{zhou, bau2020} that object detectors emerge in CNNs despite having no supervised training. Not only so, but we observe that the different channels are detectors for different kinds of objects in the input image, and that they are strongly consistent across multiple input images. For example, as shown in %Figures~\ref{fig:ImageNet} and~\ref{fig:VGG},
Figure~\ref{fig:ImageNet},
channel~6 in the first convolutional layer of ResNet 18 always 
detects (i.e., has higher activation values) the object immediately next to the class object (which in this case is \textit{Egyptian cat}). Moreover, channel~6 is the top 2 highest-ranked PCACE channel, whereas channel~24 (which consistently focuses on the main class object) ranks last in the PCACE algorithm. The magnitudes of the activations are also consistent in each channel. We visualize different PCACE-ranked channels across the figures in the paper for breadth purposes.

Not only does this show the locative power of CNNs despite being trained on only labels, but we also %make the observation
observe that the most correlated channels tend to be those that do not target the main class object, and focus on different objects in the images or the edges of the shapes in the images. Figure~\ref{fig:ImageNet} shows how some of the top 5 PCACE channels consistently focus on objects %in the image 
that are not the class object or the edges of the image shape, whereas they tend to not highlight the main class object. There are also instances when this is not the case and some of the top PCACE channels focus on the main class object, but they are scarce.

\begin{figure}[ht]
  \centering
  {\includegraphics[width=0.13\textwidth]{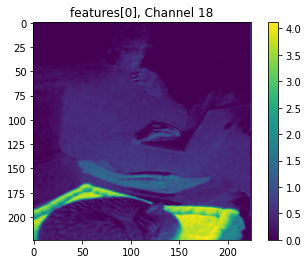}}
  {\includegraphics[width=0.13\textwidth]{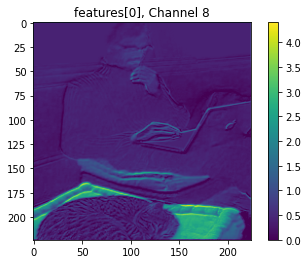}}
  {\includegraphics[width=0.13\textwidth]{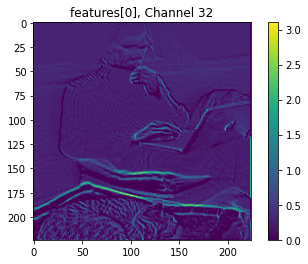}}
  \par\medskip
  {\includegraphics[width=0.13\textwidth]{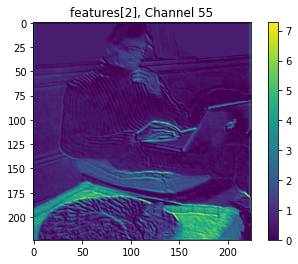}}
  {\includegraphics[width=0.13\textwidth]{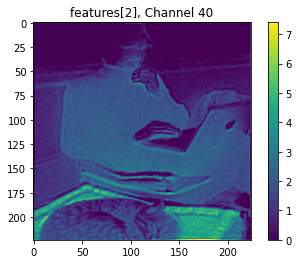}}
  {\includegraphics[width=0.13\textwidth]{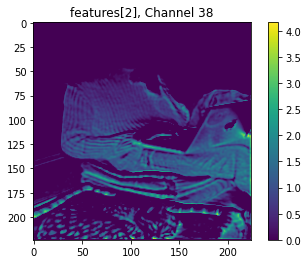}}
  \par\medskip
  {\includegraphics[width=0.13\textwidth]{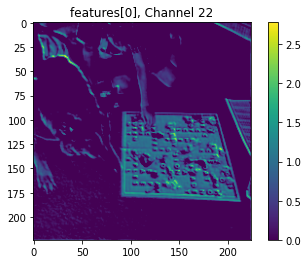}}
  {\includegraphics[width=0.13\textwidth]{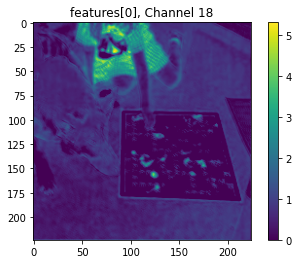}}
  {\includegraphics[width=0.13\textwidth]{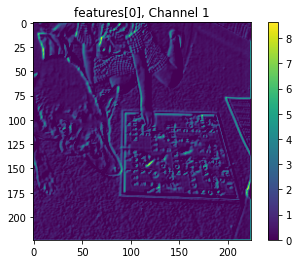}}
  \par\medskip
  {\includegraphics[width=0.13\textwidth]{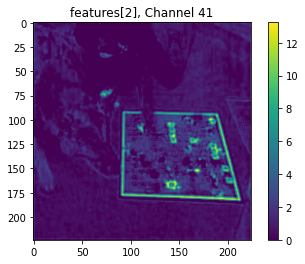}}
  {\includegraphics[width=0.13\textwidth]{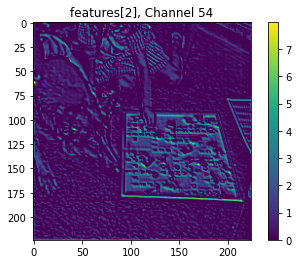}}
  {\includegraphics[width=0.13\textwidth]{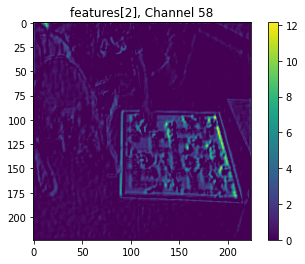}}
  \caption{Activation maps of three of the top 5 PCACE channels for %the classes 
  \textit{Egyptian cat} and \textit{German shepherd} in VGG-16.}
  \label{fig:VGG}
\end{figure}

We repeated the same visualizations with VGG-16 and obtained the same observations, which are shown in Figure~\ref{fig:VGG} (note that VGG-16 does not allow for CAM visualizations, which is why they do not appear in Figure~\ref{fig:VGG}). Therefore, these findings are not specific to the ResNet-18 architecture, and we hypothesize that they extend to other architectures as well. These observations are coherent with the findings in~\cite{li}, where they argue that different neural networks learn the same representations.

\begin{figure}[ht]
    \includegraphics[width=8.5cm]{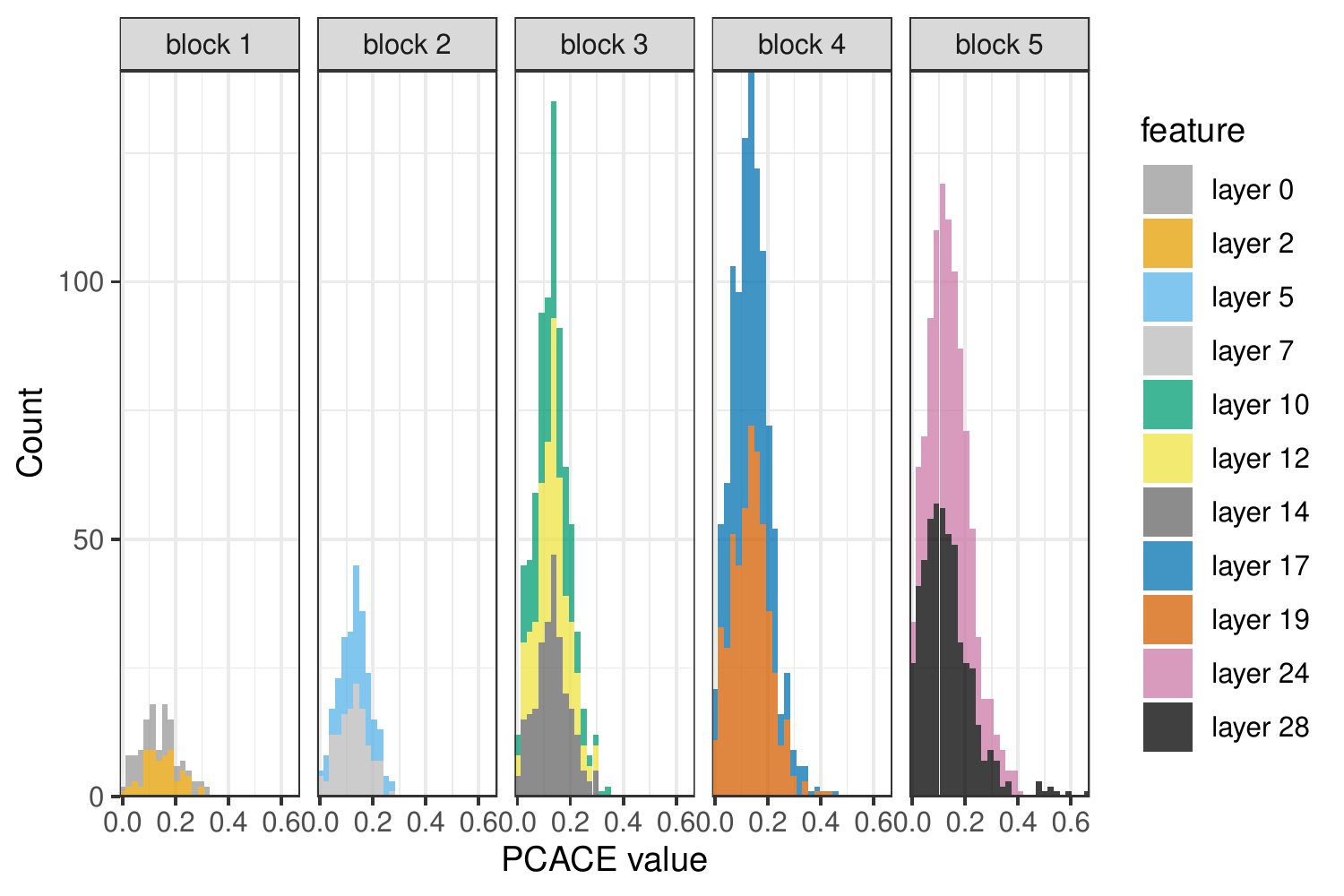}
    \caption{Histogram of the PCACE values for the successive convolutional layers in VGG-16 computed with the ImageNet class \textit{Egyptian cat}.}
    \centering
    \label{fig:sorted}
    \vspace*{-0.5cm}
\end{figure}

Finally, Figure~\ref{fig:sorted} shows sorted PCACE values for all the convolutional layers in the VGG-16 architecture for the ImageNet class \textit{Egyptian cat} in order to analyze how the correlation values vary across the different convolutional layers. 
% In Figure~\ref{fig:sorted2}, the $x$-axis indicates the numbering of the channels, which is why it varies across the network (there are 64 channels in the first two convolutional layers, 128 in the next two, 256 in the next three, and 512 in the rest). 
%We observe that 
The PCACE values follow similar trends across the different layers, and %that 
they are almost superposed. Deeper layers tend to have lower PCACE values at the smaller end, but higher correlations on the larger end (e.g., the highest PCACE value in the first Conv layer does not exceed $0.32$, whereas the highest PCACE value in Features 28 exceeds $0.65$). However, 
%the histogram in 
Figure~\ref{fig:sorted} shows that the count distribution of the PCACE values within the same block still 
differs across layers.

As %it was the case 
with MNIST, we find no relevant correlation between %the 
PCACE rankings of %the different 
ImageNet classes, hereby indicating that a global ranking cannot be directly deduced from the class-based PCACE ranking. We %also 
repeated the same experiment with the activation maximization method as we did for MNIST: we feed into the network the ideal image generated through activation maximization for each class and record the activations of each channel. We do not find any correlation between %the 
PCACE values and the maximum or mean activation of each channel, which again indicates %some discrepancies 
some discrepancy between gradient-based and correlation-based methods.

%\begin{figure}[h!]
%  \centering
%  {\hspace{0.1cm}\includegraphics[width=0.13\textwidth]{a1.jpg}}
%  \hfill
%  {\includegraphics[width=0.15\textwidth]{a2.png}}
%  {\includegraphics[width=0.15\textwidth]{a3.png}}
%  \par\bigskip
%  {\hspace{0.4cm}\includegraphics[width=0.09\textwidth]{c1.jpg}}
%  \hfill
%  {\includegraphics[width=0.15\textwidth]{c2.png}}
%  {\includegraphics[width=0.15\textwidth]{c3.png}}
%  \par\bigskip
%  {\hspace{0.1cm}\includegraphics[width=0.13\textwidth]{b1.jpg}}
%  \hfill
%  {\includegraphics[width=0.15\textwidth]{b2.png}}
%  {\includegraphics[width=0.15\textwidth]{b3.png}}
%  \par\bigskip
%  {\includegraphics[width=0.14\textwidth]{d1.jpg}}
%  \hfill
%  {\includegraphics[width=0.15\textwidth]{d2.png}}
%  {\includegraphics[width=0.15\textwidth]{d3.png}}
%  \caption{CAM visualization and the activation maps of two of the top 5 PCACE channels for the ImageNet classes \textit{Egyptian cat}, \textit{German shepherd}, and \textit{Church} in Res\-Net-18.}
%\end{figure}

%To exemplify another visualization tool that can be used with PCACE, Figure~\ref{fig:histogram} shows a histogram visualization of the 64 activation maps for each channel in the first convolutional layer of VGG-16 sorted according to the PCACE ranking applied to the ImageNet class \textit{Egyptian cat}.

\begin{figure}[ht]
    \includegraphics[width=8cm]{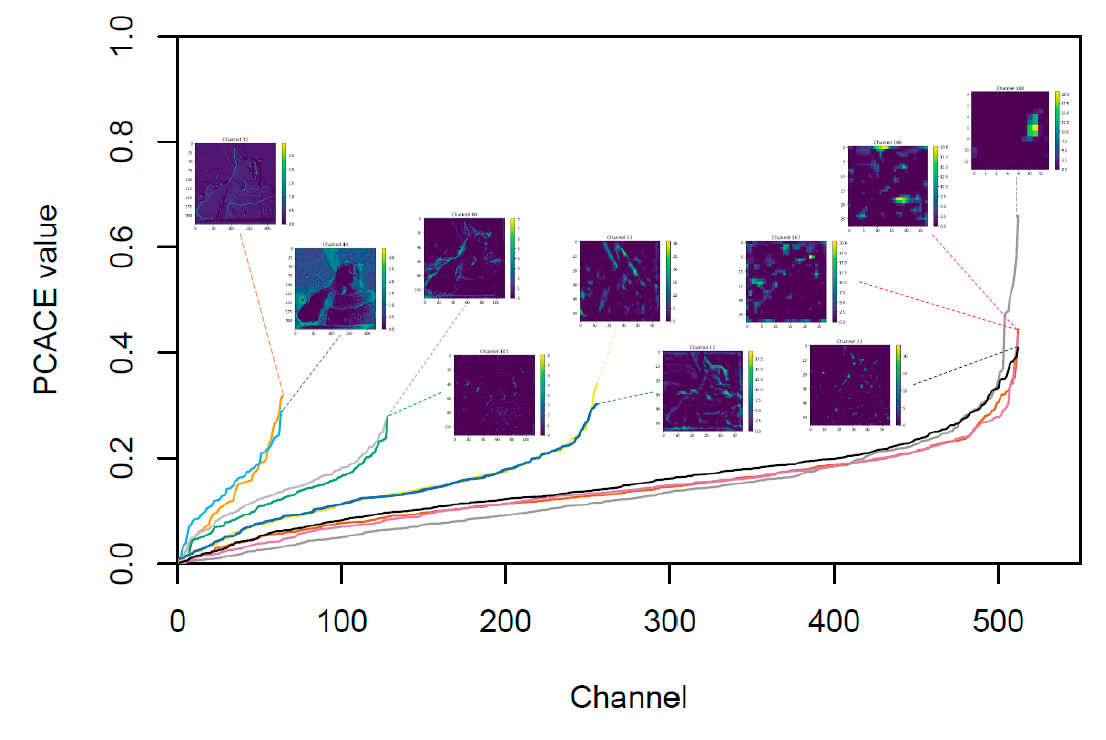}
    \caption{Sorted PCACE values for the successive convolutional layers in VGG-16 computed with the ImageNet class \textit{Egyptian cat}. Includes the activation maps of the top PCACE channel for each layer. As in Figure \ref{fig:sorted}, blocks are visible from left to right with block 1 (64 channels), block 2 (128 channels), block 3 (256 channels), and blocks 4 and 5 (overlapping, 512 channels).}
    \centering
    \label{fig:sorted2}
\end{figure}

\begin{figure*}[ht]
  \centering
  {\includegraphics[width=0.055\textwidth]{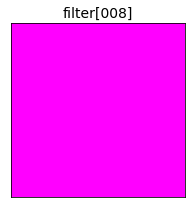}}\hspace{0.2cm}
  {\includegraphics[width=0.055\textwidth]{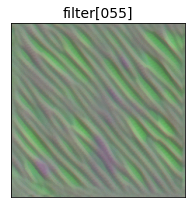}}\hspace{0.2cm}
  {\includegraphics[width=0.055\textwidth]{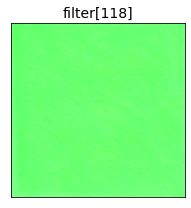}}\hspace{0.2cm}
  {\includegraphics[width=0.055\textwidth]{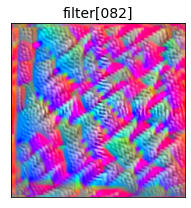}}\hspace{0.2cm}
  {\includegraphics[width=0.055\textwidth]{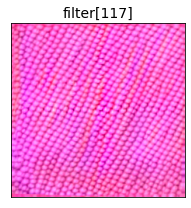}}\hspace{0.2cm}
  {\includegraphics[width=0.055\textwidth]{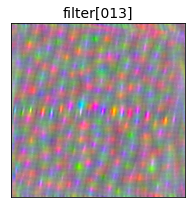}}\hspace{0.2cm}
  {\includegraphics[width=0.055\textwidth]{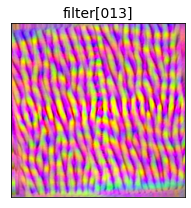}}\hspace{0.2cm}
  {\includegraphics[width=0.055\textwidth]{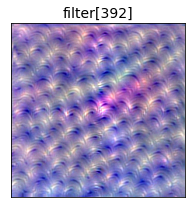}}\hspace{0.2cm}
  {\includegraphics[width=0.055\textwidth]{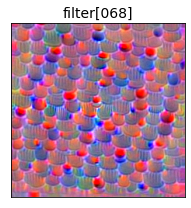}}\hspace{0.2cm}
  {\includegraphics[width=0.055\textwidth]{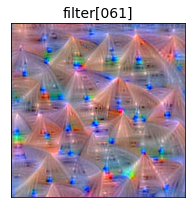}}\hspace{0.2cm}
  {\includegraphics[width=0.055\textwidth]{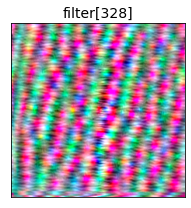}}
\par\smallskip
  {\includegraphics[width=0.055\textwidth]{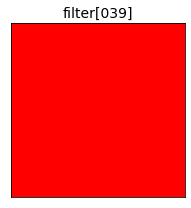}}\hspace{0.2cm}
  {\includegraphics[width=0.055\textwidth]{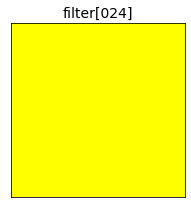}}\hspace{0.2cm}
  {\includegraphics[width=0.055\textwidth]{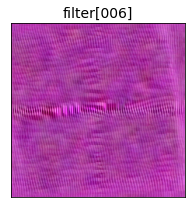}}\hspace{0.2cm}
  {\includegraphics[width=0.055\textwidth]{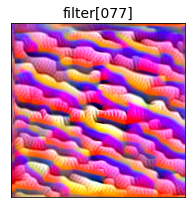}}\hspace{0.2cm}
  {\includegraphics[width=0.055\textwidth]{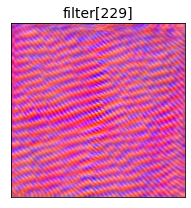}}\hspace{0.2cm}
  {\includegraphics[width=0.055\textwidth]{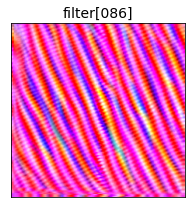}}\hspace{0.2cm}
  {\includegraphics[width=0.055\textwidth]{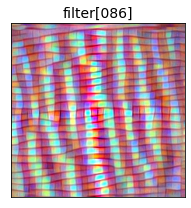}}\hspace{0.2cm}
  {\includegraphics[width=0.055\textwidth]{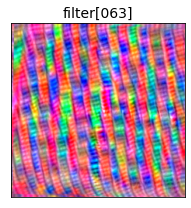}}\hspace{0.2cm}
  {\includegraphics[width=0.055\textwidth]{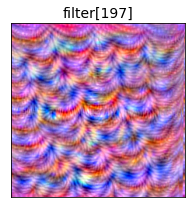}}\hspace{0.2cm}
  {\includegraphics[width=0.055\textwidth]{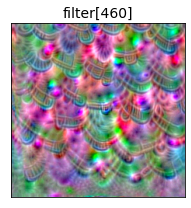}}\hspace{0.2cm}
  {\includegraphics[width=0.055\textwidth]{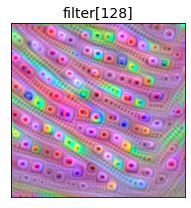}}
\par\smallskip
  {\includegraphics[width=0.055\textwidth]{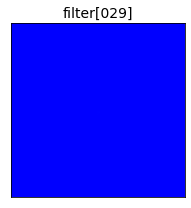}}\hspace{0.2cm}
  {\includegraphics[width=0.055\textwidth]{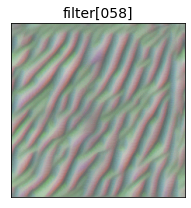}}\hspace{0.2cm}
  {\includegraphics[width=0.055\textwidth]{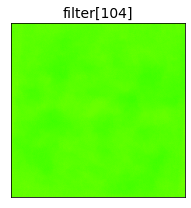}}\hspace{0.2cm}
  {\includegraphics[width=0.055\textwidth]{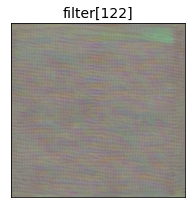}}\hspace{0.2cm}
  {\includegraphics[width=0.055\textwidth]{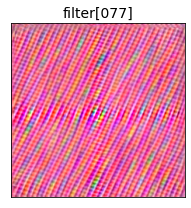}}\hspace{0.2cm}
  {\includegraphics[width=0.055\textwidth]{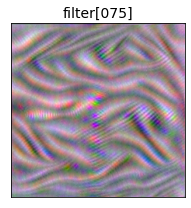}}\hspace{0.2cm}
  {\includegraphics[width=0.055\textwidth]{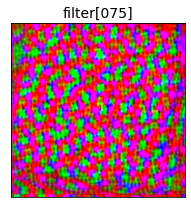}}\hspace{0.2cm}
  {\includegraphics[width=0.055\textwidth]{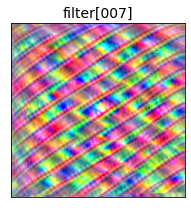}}\hspace{0.2cm}
  {\includegraphics[width=0.055\textwidth]{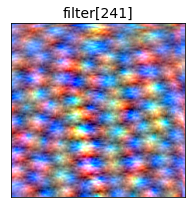}}\hspace{0.2cm}
  {\includegraphics[width=0.055\textwidth]{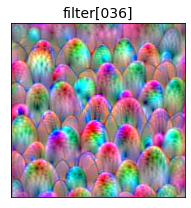}}\hspace{0.2cm}
  {\includegraphics[width=0.055\textwidth]{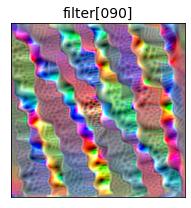}}
\par\smallskip
  {\includegraphics[width=0.055\textwidth]{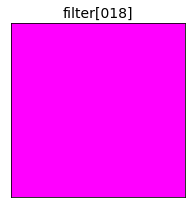}}\hspace{0.2cm}
  {\includegraphics[width=0.055\textwidth]{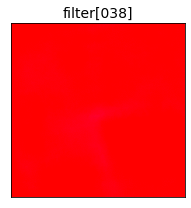}}\hspace{0.2cm}
  {\includegraphics[width=0.055\textwidth]{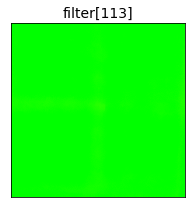}}\hspace{0.2cm}
  {\includegraphics[width=0.055\textwidth]{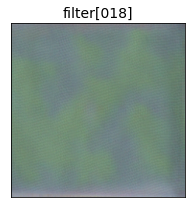}}\hspace{0.2cm}
  {\includegraphics[width=0.055\textwidth]{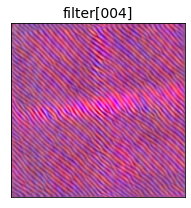}}\hspace{0.2cm}
  {\includegraphics[width=0.055\textwidth]{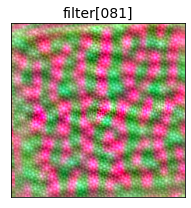}}\hspace{0.2cm}
  {\includegraphics[width=0.055\textwidth]{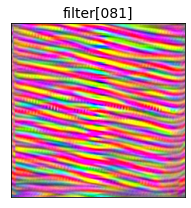}}\hspace{0.2cm}
  {\includegraphics[width=0.055\textwidth]{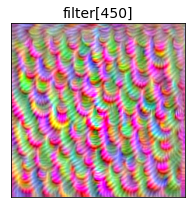}}\hspace{0.2cm}
  {\includegraphics[width=0.055\textwidth]{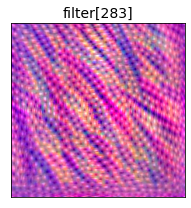}}\hspace{0.2cm}
  {\includegraphics[width=0.055\textwidth]{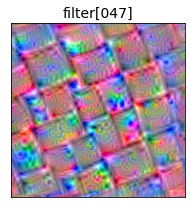}}\hspace{0.2cm}
  {\includegraphics[width=0.055\textwidth]{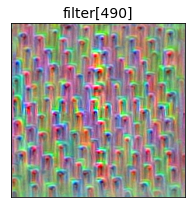}}
\par\smallskip
  {\includegraphics[width=0.055\textwidth]{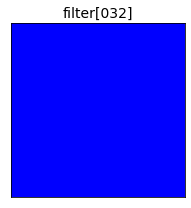}}\hspace{0.2cm}
  {\includegraphics[width=0.055\textwidth]{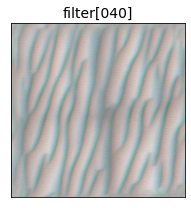}}\hspace{0.2cm}
  {\includegraphics[width=0.055\textwidth]{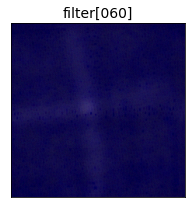}}\hspace{0.2cm}
  {\includegraphics[width=0.055\textwidth]{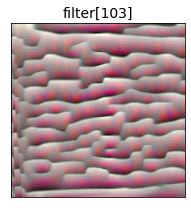}}\hspace{0.2cm}
  {\includegraphics[width=0.055\textwidth]{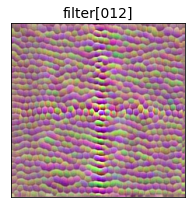}}\hspace{0.2cm}
  {\includegraphics[width=0.055\textwidth]{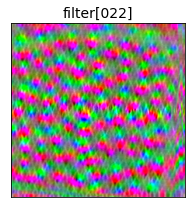}}\hspace{0.2cm}
  {\includegraphics[width=0.055\textwidth]{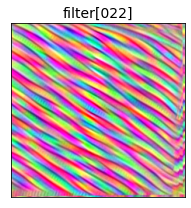}}\hspace{0.2cm}
  {\includegraphics[width=0.055\textwidth]{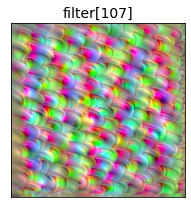}}\hspace{0.2cm}
  {\includegraphics[width=0.055\textwidth]{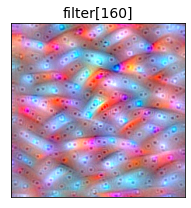}}\hspace{0.2cm}
  {\includegraphics[width=0.055\textwidth]{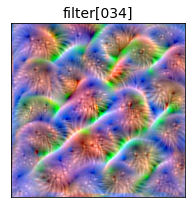}}\hspace{0.2cm}
  {\includegraphics[width=0.055\textwidth]{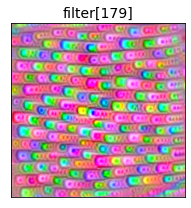}}
  \caption{Activation maximization method optimizing the top 5 PCACE channels of all convolutional layers of VGG-16 for the ImageNet class \textit{Egyptian cat} (left to right from first convolutional layers to the deeper ones).}
  \label{fig:filters}
\end{figure*}

\section{Filter Visualization with Activation Maximization}\label{sec:filter}
As explained in Section~\ref{sec:imagenet}, it becomes harder to visualize activation maps of the deeper convolutional layers as their size reduces. Therefore, we turn to the technique of activation maximization in order to analyze the feature visualizations of the different channels across multiple convolutional layers.

We produce the ideal synthesized image for each channel in all convolutional layers of VGG-16 using the library. \texttt{tf-keras-vis}\footnote{https://github.com/keisen/tf-keras-vis.} We minimize in the $\ell_2$ norm using loss steps of size 50. In Figure~\ref{fig:filters} we show the filter visualization of the top 5 PCACE channels for the class \textit{Egyptian cat} for each convolutional layer in VGG-16. Note that the ideal image of a channel is independent of the class, but the PCACE ranking is not. Like in previous situations, showing the feature visualizations for all channels would have required $4{,}224$ images. With the PCACE algorithm, we can meaningfully decide which subset to visualize without this choice being arbitrary. Figure~\ref{fig:filters} also shows how the channels are sequentially encoding more complex information as we inquire deeper into the network.

%\begin{figure}[ht]
%\includegraphics[width=8.5cm]{histogram.png}
%\caption{Histogram of the 64 channels in the first Conv layer of VGG-16 according to their PCACE values.}
%\centering
%\label{fig:histogram}
%\end{figure}

\section{Use Case: Application to Street-Level Images for Air Pollution Detection}\label{sec:airpoll}
Lastly, we show the results of the PCACE algorithm in a real-word application where the aim is to predict air pollution levels from street-level images. We use weights from a model trained by by~\cite{esra} using a slightly modified ResNet-18 architecture where the outputs are continuous pollutant levels. In this paper, we use the weights trained using a subset of images from the city of London for predicting annual $\textrm{NO}_2$ levels. 

\begin{figure*}
\centering
\begin{subfigure}{0.33\textwidth}
\centering
  {\hspace{-0.36cm}\includegraphics[width=0.9\textwidth]{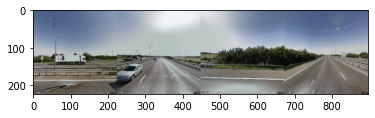}}
  {\includegraphics[width=0.8\textwidth]{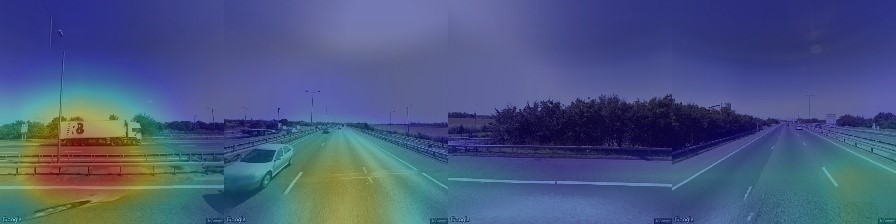}}
  \par\bigskip
 % {\includegraphics[width=0.32\textwidth]{aira3.png}}
  {\includegraphics[width=0.38\textwidth]{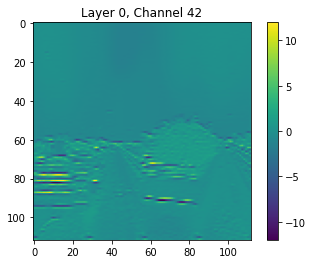}}
  {\includegraphics[width=0.38\textwidth]{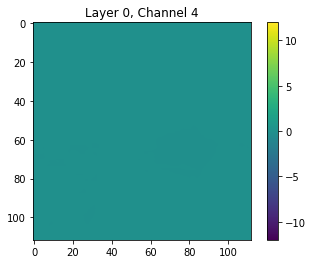}}
  \par\bigskip
  {\includegraphics[width=0.38\textwidth]{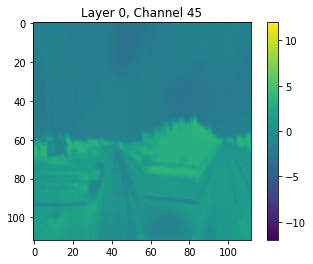}}
  {\includegraphics[width=0.38\textwidth]{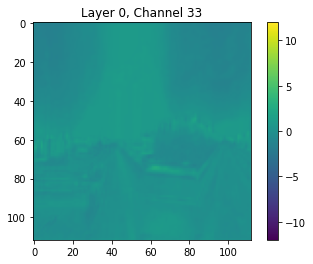}}
%\caption{Example 1}
%\label{fig:Ex1}
\end{subfigure}
\hfill
\begin{subfigure}{0.33\textwidth}
\centering
  {\hspace{-0.36cm}\includegraphics[width=0.9\textwidth]{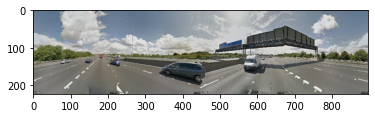}}
  {\includegraphics[width=0.8\textwidth]{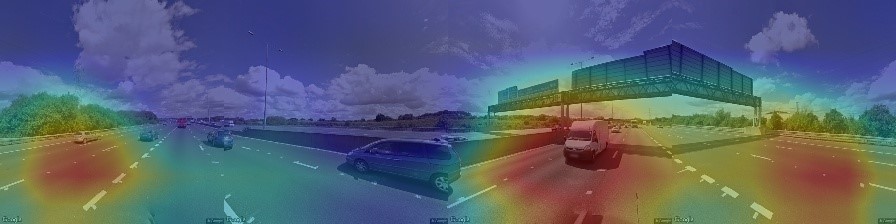}}
  \par\bigskip
  %{\includegraphics[width=0.32\textwidth]{airb3.png}}
  {\includegraphics[width=0.38\textwidth]{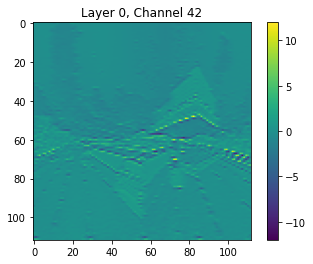}}
  {\includegraphics[width=0.38\textwidth]{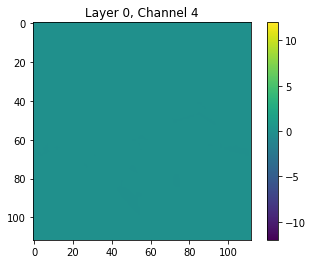}}
  \par\bigskip
  {\includegraphics[width=0.38\textwidth]{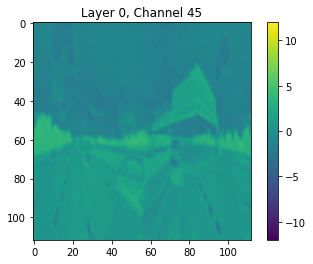}}
  {\includegraphics[width=0.38\textwidth]{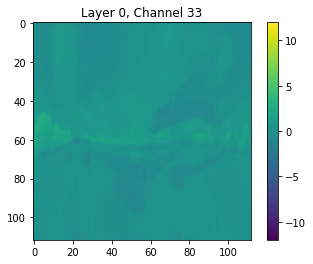}}
%\caption{Example 2}
%\label{fig:Ex2}
\end{subfigure}
\hfill
\begin{subfigure}{0.33\textwidth}
\centering
  {\hspace{-0.36cm}\includegraphics[width=0.9\textwidth]{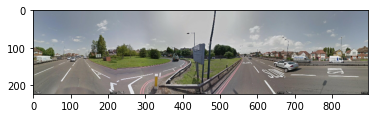}}
  {\includegraphics[width=0.8\textwidth]{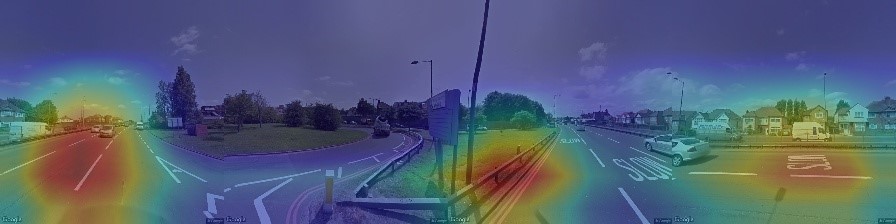}}
  \par\bigskip
  %{\includegraphics[width=0.32\textwidth]{airc3.png}}
 {\includegraphics[width=0.38\textwidth]{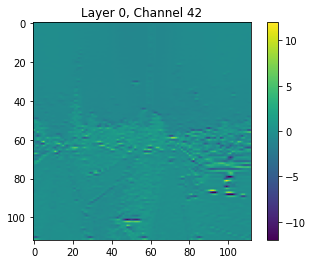}}
  {\includegraphics[width=0.38\textwidth]{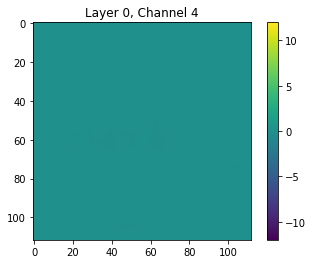}}
  \par\bigskip
  {\includegraphics[width=0.38\textwidth]{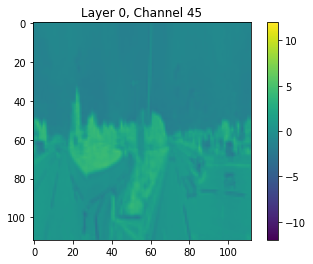}}
  {\includegraphics[width=0.38\textwidth]{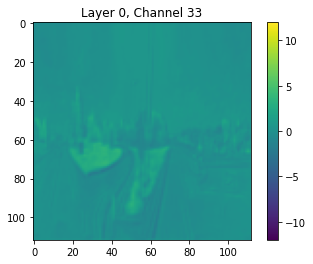}}
%\caption{Example 3}
%\label{fig:Ex3}
\end{subfigure}
\hfill
\caption{Input image (4 street-level images from a single vantage point covering 360\textdegree), CAM visualization, and activation maps of lowest 2 PCACE channels (top) and highest 2 PCACE channels (bottom) with Res\-Net-18.}
\label{fig:3Examples}
\vspace*{-0.2cm}
\end{figure*}

We include this application for two reasons: first, to present the uses of PCACE in datasets other than the typical uses of MNIST and ImageNet. Secondly, Sections~\ref{sec:mnist} and~\ref{sec:imagenet} are centered around \textit{classification tasks}. In this section, the task to predict air pollution from images corresponds to a \textit{regression task}, where there is only one final score instead of all the class scores. Still, since in this paper we have analyzed PCACE from a class-based perspective, we apply the PCACE algorithm to 300 input images from most polluted areas in the city of London (i.e., top decile). For each of these 300 images, both the true and predicted $\textrm{NO}_2$ values are above $84\;\mu \text{g}/\text{m}^3$. This prompts the following question: what are the channels detecting in high-pollution images? In %Figures~16, 17, and 18, 
Figure~\ref{fig:3Examples} we present some examples of the original input street-level image, its CAM visualization, and the activation maps of the bottom 2 and top 2 PCACE channels. 

Firstly, we show that the channels continue to act as object detectors despite having trained the architecture with a regression task instead of a classification one. Since in this setting we do not have a main class object as point of reference, understanding what the neurons are focusing on becomes a more open-ended question. We find that while some channels continue to focus on the edges of the image, many act as object detectors for buildings and, more remarkably, trees. As it was the case for MNIST and ImageNet, the activation maps of a fixed channel across different input images are consistent. For example, as can be seen in %Figures~16, 17, and 18, 
Figure~\ref{fig:3Examples}, channel 42 consistently activates at the edges of the image whereas channel 45 consistently detects the trees in the input image. 

\begin{figure*}
\centering
\begin{subfigure}{0.33\textwidth}
\centering
  {\includegraphics[width=0.65\textwidth]{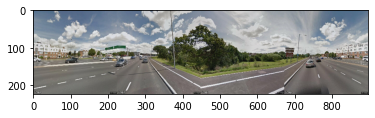}}
  {\includegraphics[width=0.33\textwidth]{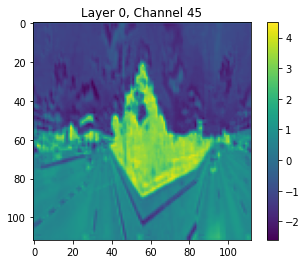}}
  \par\bigskip
  {\includegraphics[width=0.65\textwidth]{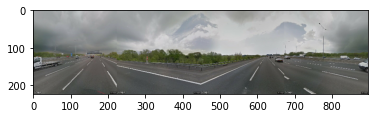}}
  {\includegraphics[width=0.33\textwidth]{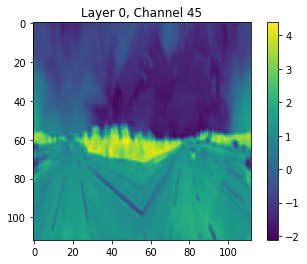}}
  \par\bigskip
  {\includegraphics[width=0.65\textwidth]{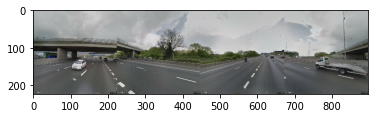}}
  {\includegraphics[width=0.33\textwidth]{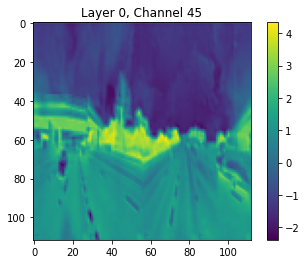}}
%\caption{}
\end{subfigure}
\hfill
\begin{subfigure}{0.33\textwidth}
\centering
  {\includegraphics[width=0.65\textwidth]{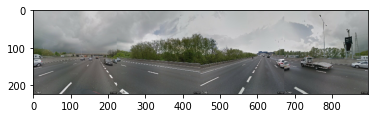}}
  {\includegraphics[width=0.33\textwidth]{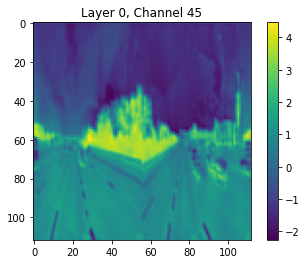}}
  \par\bigskip
  {\includegraphics[width=0.65\textwidth]{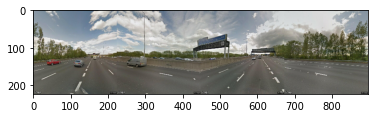}}
  {\includegraphics[width=0.33\textwidth]{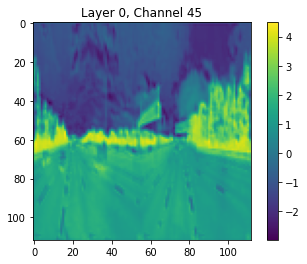}}
  \par\bigskip
  %{\includegraphics[width=0.65\textwidth]{t11.png}}
  %{\includegraphics[width=0.3\textwidth]{t12.png}}
  {\includegraphics[width=0.65\textwidth]{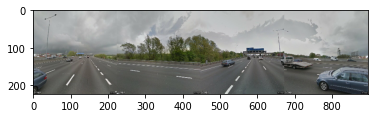}}
  {\includegraphics[width=0.33\textwidth]{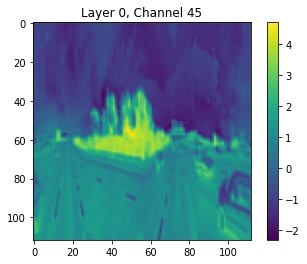}}
%\caption{}
\end{subfigure}
\hfill
\begin{subfigure}{0.33\textwidth}
\centering
  %{\includegraphics[width=0.65\textwidth]{t13.png}}
  %{\includegraphics[width=0.3\textwidth]{t14.png}}
  %\par\bigskip
  %{\includegraphics[width=0.65\textwidth]{t15.png}}
  %{\includegraphics[width=0.3\textwidth]{t16.png}}
  %\par\bigskip
  {\includegraphics[width=0.65\textwidth]{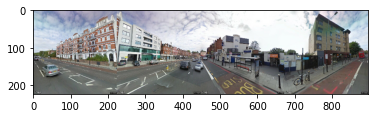}}
  {\includegraphics[width=0.33\textwidth]{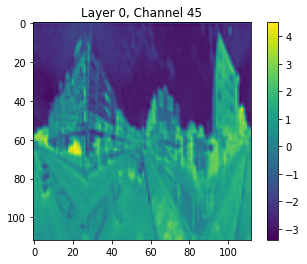}}
  \par\bigskip
  {\includegraphics[width=0.65\textwidth]{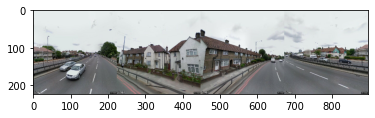}}
  {\includegraphics[width=0.33\textwidth]{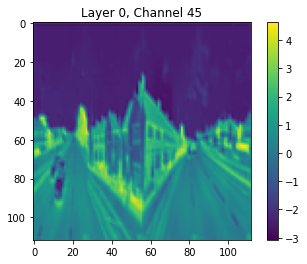}}
  \par\bigskip
  {\includegraphics[width=0.65\textwidth]{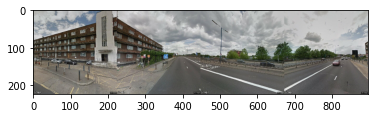}}
  {\includegraphics[width=0.33\textwidth]{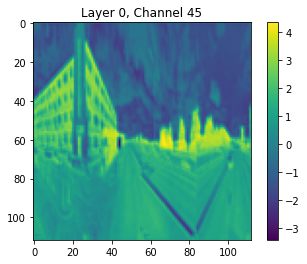}}
%\caption{}
\end{subfigure}
\hfill
%\begin{subfigure}[ht]{0.24\textwidth}
%\centering
%  {\includegraphics[width=0.65\textwidth]{t19.png}}
%  {\includegraphics[width=0.3\textwidth]{t20.png}}
%  \par\bigskip
%  {\includegraphics[width=0.65\textwidth]{t21.png}}
%  {\includegraphics[width=0.3\textwidth]{t22.png}}
%\caption{Example 4}
%\label{fig:Exx4}
%\end{subfigure}
\caption{Channel 45 in the first convolutional layer of ResNet-18 trained on ImageNet and fine-tuned with street-level images acts as a powerful tree detector. Channel 45 was the second highest scoring %channel 
using PCACE, as seen in Fig.\,\ref{fig:3Examples} (colorbar to scale).}
\label{fig:9Examples}
\end{figure*}

In particular, channel 45, which is the second highest ranked PCACE channel, is a surprisingly powerful tree detector 
(see Figure~\ref{fig:9Examples}),
%Figures 19-22), 
even when they appear as very small parts of the image, as it is the remarkable case of the %third
top right image in %Figure~21.
Figure~\ref{fig:9Examples}. More surprisingly, there is no class \textit{tree} in ImageNet $1{,}000$, which provides more evidence for the strong object detection capabilities of current CNN architectures. Moreover, we observe that the bottom PCACE channels are not doing tree detection, as shown in~\ref{fig:3Examples}. This further indicates how the PCACE ranking can help in interpreting how the network is encoding the key concepts and identifying certain classes.

When we compare the activation maps of the PCACE channels to the CAM visualizations of the same street-level images of~\cite{esra}, as we exemplify in %Figures~16, 17, and 18, 
Figure~\ref{fig:3Examples},
we observe that the CAM method tends to highlight the pixels that correspond to parts of the road, which likely indicate higher levels of pollution. As we argued, the activation maps act as object detectors, and therefore in the case of regression tasks we recommend leveraging CAM methods (which can highlight particular areas of the image beyond specific objects) with the PCACE activation maps. However, the PCACE method alone was able to rank highest the neurons that do tree object detection, and it is sensible to conclude that trees in cities are most predictive for pollutant methods, thus indicating the effectiviness of the PCACE ranking in this task.

%Nonetheless, it would be hard to extract this conclusion from the activation maps, which as we showed act as concrete object detectors, and we hypothesize are not able to highlight a particular area of an object such as \textit{road}. Therefore, in the case of regression tasks, we recommend leveraging the CAM methods with the PCACE activation maps. 

\section{Conclusions and Discussion}
In this paper we have presented the PCACE algorithm: a new statistical method that combines the Principal Component Analysis for dimensionality reduction with the Alternating Conditional Expectation algorithm to find the maximal correlation coefficient between a hidden neuron and the final class score. PCACE returns a number between 0 and 1 that indicates the strength of the non-linear relationship between the multiple predictor variables (each element in the activation matrix of a particular channel when feeding multiple input images into the network) and the response variable (each correct final class score before the softmax function when feeding multiple input images into the network). We then use the different PCACE values to rank the hidden units in order of importance. PCACE yields
%constitutes 
a rigorous statistical and, most importantly, standardized method which is able to quantify the relevance of each neuron towards %the final
classification. %We show how 
Thus, PCACE constitutes a %very 
useful tool to analyze deep learning models trained on spatiotemporal data, as shown by the top ranking of tree detector channels on our street-level air pollution imagery.

We have tested our algorithm in two well-known datasets for classification tasks, MNIST and ImageNet. We have also used it in a real-world application to street-level images for detecting air pollution, which also shows how we can use PCACE in regression tasks beyond classification ones. %Moreover, in the case of ImageNet we have tested PCACE in both the ResNet-18 and VGG-16 architectures, in order to ensure that our conclusions are not restricted to one architecture. 
We have provided extensive evidence that the channels in CNN architectures act as object detectors, and not only for the main class object in the case of classification tasks, but also for secondary objects in the images as well. Moreover, most of the top PCACE channels tend to be those that detect these secondary objects or corners and edges of the input image. We have shown that the type of object detection ability is consistent for each particular channel independent of the input image (e.g., edges, small set of pixels, secondary object). Moreover, these object detection capabilities are preserved in the case of regression tasks, as we have shown through the strong tree detection properties of the top PCACE channels in the street-level air pollution images.

We have also observed that the top ranked PCACE channels tend to not highlight the main object in their activation class, and instead focus on secondary objects of the image or on the edges of the main shapes. In the case of MNIST, we have shown how when the ideal image synthesized by the activation maximization method is fed into the architecture, the top ranked PCACE channels tend to present an activation close to 0. In the case of ImageNet, there is also no correlation between the mean or the maximum activation value in the activation map of each channel and its PCACE value. This indicates a discrepancy between gradient-based methods and correlation-based methods, and we thus recommend combining different interpretability methods that rely on both correlations and optimization methods, as they tend to provide complementary information.

The PCACE values for each convolutional layer can be used in several ways. Firstly, one can study the correlation values directly, as we did in Figures~\ref{fig:sorted} and \ref{fig:sorted2}. We showed that the PCACE values across the different convolutional layers in an architecture are coherent and follow the same trend within each block, with deeper layers acquiring higher correlation values. We believe that the coherence of the PCACE values across different convolutional layers justifies using the algorithm to compare channels between different layers. In this paper, we have always ranked the neurons within the same convolutional layer, but for future work it would be interesting to merge the PCACE values of all the neurons in the architecture given the standardized properties of PCACE.

Secondly, we can use the PCACE rankings for visualization purposes. In this paper we have focused on producing the activation maps of the top PCACE channels, as well as the synthesized image produced by the activation maximization method when optimizing for each of the channels separately. The activation maps are meant to be interpreted with respect to an input image, and hence are better for showing the object detection properties of CNNs. On the other hand, the activation maximization method allows us to perform a feature visualization study of the top PCACE channels, as shown in Figure~\ref{fig:filters}. We believe that many other visualization and interpretability methods which are channel-based can be added on top of PCACE with the goal of interpreting CNN models trained on spatiotemporal data. A pressing issue in the current research work on explainability is the infeasibility of visualizing \textit{all} of the neurons in an architecture, which range in the thousands. Previous works needed to make an arbitrary choice of which neurons to illustrate when presenting an interpretability method. The PCACE algorithm now allows to make this choice in a rigorous manner.

Other directions for future work include pruning on top of PCACE to further understand the impact of the top-ranked neurons. It would also be necessary to compare the PCACE rankings to other quantifications of the definition of \textit{importance} of a channel, as it was recently done in~\cite{bau2020}. However, there is a scarcity of work in this direction, and thus a lack of standardized methods to compare to. We hope that our algorithm provides a step towards a quantifiable notion of explainability in the deep learning field. Lastly, we could consider to perform a non-class-based PCACE, and run the algorithm with a weighted set of images from all classes. We could also aim to incorporate the possible correlations between pairs of neurons, or even consider computing the PCACE with groups of neurons as predictor variables, as they have also been found to work in groups (\cite{fong2}, \cite{nguyen}). There is still much to be done to open up these black boxes.

\newpage

\section{Supplementary Material}
We provide pseudocode for the PCACE algorithm in Algorithm~\ref{pcace}. For reproducibility purposes, the relevant code and data can be found at \url{https://github.com/silviacasac/ranking-CNN-neurons}.

\begin{algorithm}
\caption{PCACE Algorithm.}\label{pcace}
\begin{algorithmic}[1]
\State \emph{For each channel $c$ in the conv layer $l$ do}:
\State \hspace{0.3cm}\emph{For each input image do}:
\State \hspace{0.6cm}\emph{Store the activations of channel $c$ into matrix $X_{c, l}$.}
\State \hspace{0.6cm}\emph{Store the final class score into vector $Y_{c, l}$.}
\State \hspace{0.3cm}\emph{Apply PCA to reduce the dimensionality of matrix $X_{c, l}$.}
\State \hspace{0.3cm}\emph{Obtain newly reduced matrix $X'_{c, l} = \{X'_1, X'_2, \ldots, X'_p\}$.}
\State \hspace{0.3cm}\emph{Let $\Theta(Y_{c, l}), \phi_1(X'_1), \ldots, \phi_p(X'_p)$ be zero-mean functions.}
\State \hspace{0.3cm}\emph{Let $e^2(\Theta, \phi_1, \ldots, \phi_p) = \frac{\mathbb{E}[\Theta(Y_{c, l}) - \sum_{i=1}^p \phi_i(X'_i)^2]}{\mathbb{E}[\Theta^2(Y_{c, l})]}$.}
\State \hspace{0.3cm}\emph{Let $T$ be the error tolerance}.
\State \hspace{0.3cm}\emph{While $e^2 > T$, do:}
\State \hspace{0.6cm}\emph{Holding $\phi_1(X'_1), \ldots,\phi_p(X'_p)$ fixed, minimizing $e^2$ yields}
\State \hspace{2.2cm}\emph{$\Theta_1(Y_{c, l}) = \mathbb{E}[\sum_{i=1}^p \phi_i(X'_i) \mid Y_{c, l}]$.}
\State \hspace{0.6cm}\emph{Normalize $\Theta_1(Y_{c, l})$ to unit variance.}
\State \hspace{0.6cm}\emph{For each $k$, fix other $\phi_i(X'_i)$ and $\Theta(Y_{c, l})$ minimizing $e^2$.}
\State \hspace{0.6cm}\emph{The solution is $\Tilde{\phi_k} = \mathbb{E} [\Theta(Y_{c, l}) - \sum_{i \neq k} \phi_i(X'_i) \mid X'_k]$.}
\State \hspace{0.3cm}\emph{Return the absolute value of the Pearson product-moment}
\State \hspace{0.3cm}\emph{correlation coefficients.}
\State \emph{Sort the final $c$ values.}
\end{algorithmic}
\end{algorithm}

%\begin{algorithm}
%\caption{PCACE Algorithm}\label{pcace}
%\begin{algorithmic}[1]
%\BState \emph{For each channel $c$ in the Conv layer do}:
%\BState \emph{For each input image do}:
%\BState \emph{Store the activations of channel $c$ into matrix $X$.}
%\BState \emph{Store the final class score into vector $Y$.}
%\BState \emph{Apply PCA to reduce the dimensionality of matrix $X$.}
%\BState \emph{Apply ACE to compute the maximal correlation coefficient between the newly reduced matrix $X'$ and vector $Y$.}
%\BState \emph{Take absolute value and sort them.}
%\end{algorithmic}
%\end{algorithm}

\end{document}